\documentclass[10pt,twocolumn,letterpaper]{article}

\usepackage{cvpr}              

\usepackage[dvipsnames]{xcolor}

\usepackage{diagbox}
\usepackage{array}
\usepackage{makecell}
\newcolumntype{C}[1]{>{\centering\arraybackslash}m{#1}}

\definecolor{cvprblue}{rgb}{0.21,0.49,0.74}
\usepackage[pagebackref,breaklinks,colorlinks,citecolor=cvprblue]{hyperref}

\def\paperID{4} 
\def\confName{EVAL-FoMo 24}
\def\confYear{2024}

\title{Can Visual Language Models Replace OCR-Based Visual Question Answering Pipelines in Production? A Case Study in Retail.}

\author{
Bianca Lamm\\
Markant Services International GmbH\\
Offenburg, Germany\\
{\tt\small Bianca.Lamm@de.markant.com}
\and
Janis Keuper\\
Institute for Machine Learning and Analytics (IMLA)\\
Offenburg University, Germany \\
{\tt\small keuper@imla.ai}
}

\begin{document}
\maketitle
\begin{abstract}
\noindent Most production-level deployments for Visual Question Answering (VQA) tasks are still build as processing pipelines of independent steps including image pre-processing, object- and text detection, Optical Character Recognition (OCR) and (mostly supervised) object classification. However, the recent advances in vision Foundation Models \cite{zhou2023comprehensive} and Vision Language Models (VLMs) \cite{zhang2024vision} raise the question if these custom trained, multi-step approaches can be replaced with pre-trained, single-step VLMs.\\ 
This paper analyzes the performance and limits of various VLMs in the context of VQA and OCR \cite{liu2023improved, hong2023cogagent, chen2023internvl} tasks in a production-level scenario. Using data from the \textit{Retail-786k} \cite{lamm2024retail786k} dataset, we investigate the capabilities of pre-trained VLMs to answer detailed questions about advertised products in images. Our study includes two commercial models, GPT-4V \cite{OpenAI2023} and GPT-4o \cite{OpenAI2024}, as well as four open-source models: InternVL \cite{chen2023internvl}, LLaVA 1.5 \cite{liu2023improved}, LLaVA-NeXT \cite{liu2024llavanext}, and CogAgent \cite{hong2023cogagent}.\\
Our initial results show, that there is in general no big performance gap between open-source and commercial models. However, we observe a strong task dependent variance in VLM performance: while most models are able to answer questions regarding the product brand and price with high accuracy, they completely fail at the same time to correctly identity the specific product name or discount. This indicates the problem of VLMs to solve fine-grained classification tasks as well to model the more abstract concept of discounts.
\end{abstract}
\section{Introduction}
\label{sec:intro}
\begin{figure}[t!]
    \centering
    \includegraphics[width=\columnwidth]{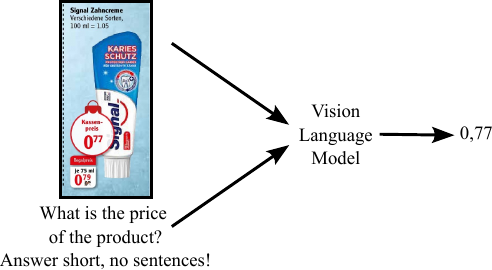}
    \caption{Illustration of the single-step process: input, model, output. The input consists of a product advertising image and a prompt querying specific product or advertising feature. A VLM is used as a model. }
    \label{fig:visual_abstract}
\end{figure}
Advances in the field of Computer Vision have been remarkable in recent years. The development of Large Language Models (LLMs) \cite{zhao2023survey}
and Vision Language Models (VLMs) \cite{zhang2024vision} that can analyze and interpret text and in the second case also images has played a significant role in this progress. The importance of handling multi-modal inputs is highlighted by the growing use of image analysis and image creation.
Previous research has shown that VLMs can be effective in Visual Question Answering (VQA), Optical Character Recognition (OCR), or Image Captioning \cite{liu2023improved, hong2023cogagent, chen2023internvl}. This study examines the transformation of a multi-step approach into a single-step process through the utilization of VLMs. The considered problem includes an OCR-based pipeline. Hence, the research question arises: \textit{can we replace OCR-based VQA pipelines with VLMs at a production level?} We investigate this question on a use case derived from the retail domain. The basis of the case study is the dataset \textit{Retail-786k} \cite{lamm2024retail786k} that consists of images cropped from leaflets. 
Each image presents an advertisement of a product. Such an advertisement contains information about the promotion and the product itself. A frequent query pertaining to the images is related to the price at which the product is being promoted. The objective of our study is the extraction of product and/or advertisement information from the images. This task refers to VQA \cite{VQA}. Initially, the task can be approached through a multi-step process: first, the detection of the image part that illustrates the information, e.g. the price information, and afterwards, extract the text by OCR. We investigate whether it is possible to obtain the desired information solely using VLMs. Figure \ref{fig:visual_abstract} illustrates the process of our study. 
The question according to the product and/or advertisement information, serving as prompt, and the image are inputs to the VLMs. Subsequently, the models return a prediction for the request.
\section{Related Work}
\label{sec:related_work}

\paragraph{Dataset.}
\label{subsec:relwork_dataset}
The dataset \textit{Retail-786k} \cite{lamm2024retail786k} contains 786,179 images that are divided into 3,298 classes. The images of the dataset were extracted from scanned advertising leaflets collected over multiple years from various European retailers. Each image comprises three distinct elements: the product image itself, the product information (e.g., name, brand, weight), and the advertisement information (e.g., promoted price, striketrough price, and discount). The dataset is provided with a longer image edge size of 256 and 512, respectively. The experiments in Section \ref{sec:experiments} are based on the longer image edge fixed at 512.

\paragraph{LLMs and VLMs.}
\label{subsec:relwork_VLMs}
LLMs and VLMs are AI systems being created to process and generate text.
VLMs can process visual and textual data to enhance understanding and generation of multi-modal content.
One of the famous LLMs is \texttt{ChatGPT}, which belongs to the series of GPT models \cite{OpenAI_GPT} of OpenAI.
The company also provides VLMs such as \texttt{GPT-4V} \cite{OpenAI2023} and \texttt{GPT-4o} \cite{OpenAI2024}.
In addition, there are open-source VLMs like \texttt{InternVL} \cite{chen2023internvl}, \texttt{LLaVA 1.5} \cite{liu2023improved}, \texttt{LLaVA-NeXT} \cite{liu2024llavanext}, or \texttt{CogAgent} \cite{hong2023cogagent}.

\paragraph{VQA.}
\label{subsec:relwork_VQA}
The objective of the VQA task is to elicit a response in the form of a verbal answer to a given image, accompanied by a question pertaining to the image itself. The question and the answer can be open-ended / non-binary. To solve the VQA task, a detailed understanding of the image is necessary \cite{VQA}.
Previous approaches to the VQA problem have been based on neural networks or Deep Learning models \cite{wu2017visual, de2023visual}. But also including architectures such as Knowledge Graph or transformer are used to develop new processes to solve VQA \cite{de2023visual}. The authors of \cite{de2023visual} describe a survey of previous and novel techniques. Current approaches also use the architectures of LLMs and VLMs \cite{wang2023towards}. In addition, training models with domain-specific knowledge can improve the VQA result in a given dataset \cite{wang2023fashionvqa}.
\section{Materials and Methods}
\label{sec:material_methods}

\paragraph{Models and Architectures.}
\label{subsec:exp_models_architectures}

All experiments in Section \ref{sec:experiments} are conducted by using cloud-based computational resources.
In practice, the cloud Microsoft Azure \cite{AzureCloud2024} is used.
Our focus is on the investigations of VLMs.
Hence, we examine the models: \texttt{GPT-4V} \cite{OpenAI2023} and the \texttt{GPT-4o} \cite{OpenAI2024} from OpenAI and further specific open-source models that were introduced subsequently.\\
Requests using the GPT models (\texttt{GPT-3.5 Turbo}, \texttt{GPT-4V}, \texttt{GPT-4o}) are executed using a CPU.
For the open-source models \texttt{InternVL} \cite{chen2023internvl}, \texttt{LLaVA 1.5} \cite{liu2023improved}, \texttt{LLaVA-NeXT} \cite{liu2024llavanext}, and \texttt{CogAgent} \cite{hong2023cogagent}, a GPU A100 is used.
The model with the longest duration among all open-source models requires about 20 minutes to query the amount of prompts. However, the models, \texttt{GPT-4V} and \texttt{GPT-4o}, each require around 2.5 hours.



\paragraph{Data and Prompt.}
\label{subsec:exp_data}
For the experiments, 50 images were randomly sampled from the dataset, ensuring that they were assigned to different classes. In addition, attention was paid to ensure that diverse products were represented in the advertisements and that the images were sourced from various retailers.
The syntax of the used prompt is: \texttt{"What is the [feature] of the product? Answer short, no sentences!"}. The values of the term feature are: "name", "brand", "weight", "price", "strikethrough price", "discount", "reference weight", and "price of the reference weight". The model \texttt{GPT-4o} has an additional term for the prompt: \texttt{"What is the [feature] of the product? Answer short, no sentences! If the information is not available, return None."}.



\paragraph{Method.}
\label{subsec:method}
A total of 400 prompt queries per model were issued during the course of the study. Due to a subset of 50 images and requesting for 8 features. It should be noted that the presence of the requested feature is not guaranteed in each image. The used dataset comprises a total of 245 distinct features that are observable in the images. The queries were conducted for each model, and subsequently, a manual comparison was undertaken between the generated predictions and the available Ground Truth (GT).
\section{Experiments}
\label{sec:experiments}

\subsection{Text Only Evaluation}
\label{subsec:exp_text_evaluation}
\begin{figure}[t!]
    \centering
    \includegraphics[width=\columnwidth]{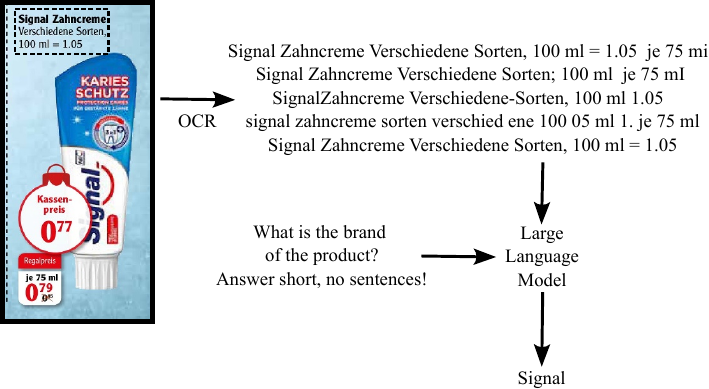}
    \caption{Illustration of the process with the use of an LLM. The prompt for the LLM is formed by the extracted text from the image getting by different OCR tools and the question about the feature "brand" of the product.}
    \label{fig:process_llm}
\end{figure}

For the investigation of an LLM, we use the model \texttt{GPT-3.5 Turbo} with the specific model id \texttt{gpt-35-turbo-16k} \cite{Azure_Models}.
The LLM model \texttt{GPT-3.5 Turbo} uses only text as input.
Therefore, the text from the images must be provided.
We use a YOLOv7-based detection model \cite{wang2023yolov7} to get the bounding box of the text in the images and then extract the text using several OCR tools.
We use the following OCR tools:
PyTesseract \cite{Hoffstaetter2014pytesseract}, EasyOCR \cite{JaidedAI2020EasyOCR}, docTR \cite{doctr2021}, OFA-OCR \cite{ofa_ocr}, and surya \cite{surya2024}.
The extracted text of all OCR tools plus the question about the feature is given to the LLM as a prompt.
The system description of the LLM is described by: \texttt{You are an AI assistant that helps people find information. You get outputs of OCR tools in which you should find the answer.} The LLM prompt only considers the feature "brand". This feature of the product is always written in the text of the image. Table \ref{tab:prompt_gpt35} shows an example of a prompt of the LLM \texttt{GPT-3.5 Turbo}. The first part of the prompt consists of the outputs of the different OCR tools. Afterwards, the request to the specific feature "brand" follows. In Figure \ref{fig:process_llm} the total LLM process is illustrated.
\begin{table}
  \caption{Illustration of an example prompt to the LLM \texttt{GPT-3.5 Turbo}.}
  \label{tab:prompt_gpt35}
  \centering
  \begin{tabular}{lp{5.5cm}}
    \toprule
                    &prompt\\
    \midrule
    PyTesseract     &\it{''Signal Zahncreme Verschiedene Sorten, 100 ml = 1.05  je 75 mi''}\\
    EasyOCR         &\it{''Signal Zahncreme Verschiedene Sorten; 100 ml  je 75 mI''}\\
    docTR           &\it{''SignalZahncreme Verschiedene-Sorten, 100 ml 1.05''}\\
    OFA-OCR         &\it{''signal zahncreme sorten verschied ene 100 05 ml 1. je 75 ml''}\\
    surya           &\it{''Signal Zahncreme Verschiedene Sorten, 100 ml = 1.05''}\\
    \midrule
    feature         &What is the brand of the product? Answer short, no sentences!\\
  \bottomrule
  \end{tabular}
\end{table}

\begin{table}[!b]
  \caption{Examples of predictions from the \texttt{GPT-4o} model that are considered as acceptable.}
  \label{tab:prediction_acceptable}
  \centering
  \begin{tabular}{@{}p{1.3cm}p{2.5cm}l@{}}
    \toprule
    feature    &GT     &\texttt{GPT-4o}\\
    \midrule
    brand       &\it{''Benediktiner''}       &\it{''Benediktiner Weissbräu''}\\
    price       &\textit{"16.99"}              &\textit{"16,99 Euro"}\\
    weight      &\textit{"200 Gramm"}          &\textit{"200 g"}\\
  \bottomrule
  \end{tabular}
\end{table}
\subsection{Image Evaluation}
\label{subsec:exp_evaluation}
\begin{figure*}[!h]
    \centering
    \begin{subfigure}[b]{0.3\textwidth}
        \includegraphics[width=0.8\textwidth]{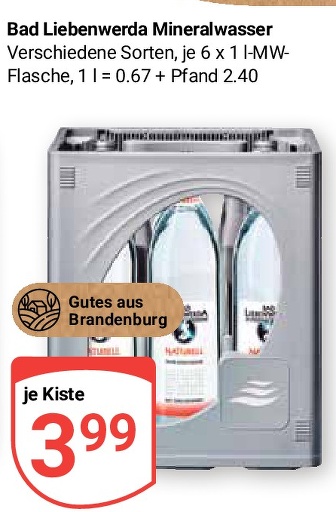}
        \caption{}
        \label{fig:image_failed_Mineralwasser}
    \end{subfigure}
    \hfill
    \begin{subfigure}[b]{0.3\textwidth}
        \includegraphics[width=\textwidth]{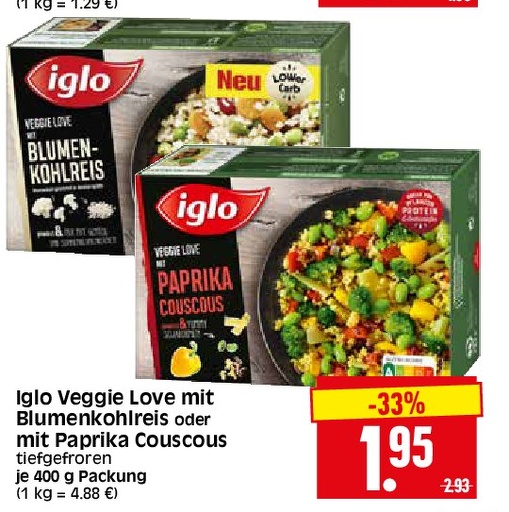}
        \caption{}
        \label{fig:image_failed_Iglo}
    \end{subfigure}
    \hfill
    \begin{subfigure}[b]{0.3\textwidth}
        \includegraphics[width=\textwidth]{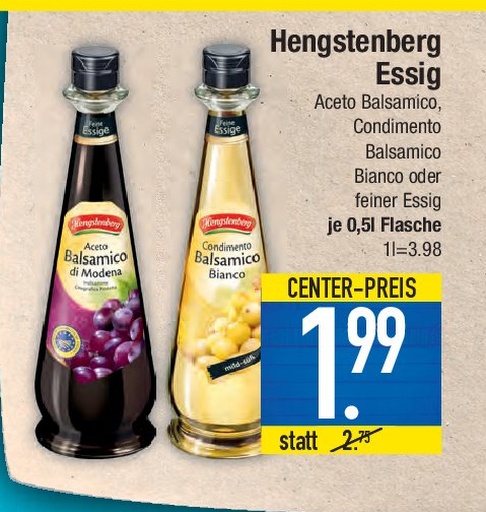}
        \caption{}
        \label{fig:image_failed_Essig}
    \end{subfigure}
    \caption{Illustration of the images for which the models produce false predictions that are illustrated in Table \ref{tab:examples_images_fail}.}
    \label{fig:examples_images_fail}
\end{figure*}

For the investigations of different VLMs, we focus on the commercial models \texttt{GPT-4V} \cite{OpenAI2023} and \texttt{GPT-4o} \cite{OpenAI2024} from OpenAI. The open-source models provide different versions of models depending on the model size. We use the following specific open-source models: \texttt{internvl\_chat\_1\_2\_plus} \cite{chen2023internvl}, \texttt{llava\_1\_5\_7b} \cite{liu2023improved}, \texttt{llava\_1\_6\_34b} \cite{liu2024llavanext}, and \texttt{cogagent\_chat} \cite{hong2023cogagent}. The predictions of the models were manually evaluated. The criteria for their acceptability requires not the exact match in case of string similarity with the GT. Table \ref{tab:prediction_acceptable} shows examples of predictions that are seem as acceptable.
Table \ref{tab:prediction_of_feature_per_model} illustrates for each feature the ratio of acceptable predictions per model.
\begin{table*}[!t]
  \caption{Illustration of the ratio of the acceptable predictions of each feature per model.}
  \label{tab:prediction_of_feature_per_model}
  \centering
  \begin{tabular}{p{2.2cm}cccccccc}
    \toprule
    \diagbox{feature}{model}
    &\makecell[c]{\texttt{OCR + }\\\texttt{GPT3.5}} 
    &\makecell[c]{\texttt{Intern}\\\texttt{VL}}
    &\makecell[c]{\texttt{LLaVA}\\\texttt{1.5}}
    &\makecell[c]{\texttt{LLaVA-}\\\texttt{NeXT}}
    &\makecell[c]{\texttt{Cog}\\\texttt{Agent}}
    &\makecell[c]{\texttt{GPT-}\\\texttt{4V}}
    &\makecell[c]{\texttt{GPT-}\\\texttt{4o}}
    &\texttt{[$\min, \max$]}\\
    \midrule
    name                        
    &-      &0.0    &\textbf{0.02}       &0.0        &0.0        &\textbf{0.02}       &0.0  &[0.0, 0.02]\\
    brand                       
    &0.84   &0.88   &0.56       &0.82       &0.92         &0.98     &\textbf{1.0}   &[0.56, 1.0]\\
    weight                      
    &-      &0.32   &0.24       &0.3        &0.3        &\textbf{0.56}       &0.46  &[0.24, 0.56]\\
    price                       
    &-      &0.9       &0.58       &0.88       &0.94         &0.98     &\textbf{1.0}    &[0.58, 1.0]\\
    \makecell[l]{strikethrough\\price} 
    &-      &0.3    &0.0        &0.36       &0.16       &0.4        &\textbf{0.88}  &[0.0, 0.88]\\
    discount                    
    &-      &0.42   &0.14       &0.38       &0.4        &0.42       &\textbf{0.72}  &[0.14, 0.72]\\
    \makecell[l]{reference\\weight}            
    &-      &\textbf{0.3}    &0.12       &0.12       &0.24       &0.08       &0.08  &[0.08, 0.3]\\
    \makecell[l]{price of the ref-\\erence weight}   
    &-      &0.08   &0.0        &0.16       &0.04       &\textbf{0.5}        &0.46  &[0.0, 0.5]\\
  \bottomrule
  \end{tabular}
\end{table*}
Despite the fact that the LLM \texttt{GPT-3.5 Turbo} is only applied to the feature "brand", the model is able to generate usable predictions with precision, specifically for 42 of the 50 images. It is regrettable that this model is not applicable to the other features, as the requisite information is not (always) available in the extracted text.
The results of the \texttt{GPT-4o} model are outstanding, with all predictions for the feature "brand" for each image considered usable. The model \texttt{GPT-4V} exhibited only a failure on a single image.
Table \ref{tab:prediction_of_feature_per_model} also shows that the models \texttt{GPT-4V} and \texttt{GPT-4o} exhibit the most acceptable predictions for the feature "price". However, the models \texttt{InternVL} and \texttt{CogAgent} also yield results of satisfactory quality.
It is of particular importance to note that the emergence of transposed digits can have a significant (negative) impact on predictions for the feature "price". Especially, if there is no further manually inspection of the predictions in a production pipeline.
With regard to the features, the "strikethrough price" and the "discount", none of the VLMs were found to be particularly convincing.
A further conclusion that can be drawn from Table \ref{tab:prediction_of_feature_per_model} is that all VLMs fail to achieve acceptable predictions for the features "reference weight" and "price of the reference weight".
It is assumed that the VLMs are unaware of the specific meanings attributed to aforementioned features. Furthermore, the information related to these features is consistently presented in a small font size in the images.
Table \ref{tab:examples_images_fail} shows the results of three images that are illustrated in Figure \ref{fig:examples_images_fail}. The requested features, the GT and the predictions of the models are indicated for each image. This investigation focuses on the four models \texttt{InternVL}, \texttt{CogAgent}, \texttt{GPT-4V}, \texttt{GPT-4o}, as these perform best on average across all features. The outcome of Table \ref{tab:examples_images_fail} demonstrate that the models do not include information written in a small font size as the features "reference weight" and the "price of the reference weight".
\begin{table*}[!h]
  \caption{Examples of predictions of images illustrated in Figure \ref{fig:examples_images_fail}. The four considered models fail on all examples.}
  \label{tab:examples_images_fail}
  \centering
  \begin{tabular}{ccccccc}
    \toprule
    image   &feature    &GT   &\texttt{InternVL}    &\texttt{CogAgent}   &\texttt{GPT-4V} &\texttt{GPT-4o}\\
    \midrule
    Figure \ref{fig:image_failed_Mineralwasser} 
    &price of the reference weight
    &\textit{"0.67"}   &\textit{"3.99"}   &\textit{"3.99"}   &\textit{"3.99"}   &\textit{"3.99"}\\
    Figure \ref{fig:image_failed_Iglo} 
    &reference weight
    &\textit{"1 Kilogramm"}    &\textit{"400 g"}  &\textit{"1.29"}   &\textit{"400 g"}  &\textit{"400 g"}\\
    Figure \ref{fig:image_failed_Essig} 
    &strikethrough price
    &\textit{"2.75"}   &\textit{"2.99"}   &\textit{"1.99"}   &\textit{"2.79"}   &\textit{"2.79"}\\    
  \bottomrule
  \end{tabular}
\end{table*}
\section{Conclusion}
\label{sec:conclusion}
The objective of this study is to examine the potential of using VLMs in the VQA task with images from the \textit{Retail-786k} dataset.
A number of commercial and open-source models are subjected to investigation.
The goal of the VQA task was to obtain specific product and advertising information from the images.
Specific product information such as the "brand" are predicted with a high quality in our test set. 
This appearance does not only refer to commercial models, the open-source VLMs achieve acceptable results, too.
It is acknowledged that all VLMs inherent difficulties in accurately predicting features such as "strikethrough price" or "reference weight" of the promoted product. It is assumed that the models have a lack domain-specific knowledge regarding the questioned features. For further investigation, the utilization of the Retrieval Augmented Generation (RAG) \cite{gao2023retrieval} method is worth considering. With this method, domain-specific knowledge can be provided to VLMs.
\newpage
{
    \small
    \bibliographystyle{ieeenat_fullname}
    \bibliography{main}
}


\end{document}